\documentclass{article}
\usepackage{spconf,amsmath,graphicx,hyperref}
\usepackage{xcolor}
\usepackage{booktabs}
\usepackage{multirow}
\usepackage{makecell}
\usepackage{comment}

\title{Multi-Turn Physics-Informed Vision-Language Model for \\Physics-Grounded Anomaly Detection\thanks{Accepted by IEEE ICASSP2026.}}
\name{Yao Gu$^{\star}$ \qquad Xiaohao Xu$^{\dagger}$ \qquad Yingna Wu$^{\star}$}
\address{$^{\star}$Shanghaitech University \quad $^{\dagger}$University of Michigan, Ann Arbor}
\begin{document}
%
\maketitle
\begin{abstract}
Vision-Language Models (VLMs) demonstrate strong general-purpose reasoning but remain limited in physics-grounded anomaly detection, where causal understanding of dynamics is essential. Existing VLMs, trained predominantly on appearance-centric correlations, fail to capture kinematic constraints, leading to poor performance on anomalies such as irregular rotations or violated mechanical motions. We introduce a physics-informed instruction tuning framework that explicitly encodes object properties, motion paradigms, and dynamic constraints into structured prompts. By delivering these physical priors through multi-turn dialogues, our method decomposes causal reasoning into incremental steps, enabling robust internal representations of normal and abnormal dynamics. Evaluated on the Phys-AD benchmark, our approach achieves 96.7\% AUROC in video-level detection—substantially outperforming prior SOTA (66.9\%)—and yields superior causal explanations (0.777 LLM score). This work highlights how structured physics priors can transform VLMs into reliable detectors of dynamic anomalies.
\end{abstract}

\section{Introduction}
\label{sec:intro}
Anomaly detection in industrial and mechanical systems is a long-standing problem in computer vision. Traditional methods focus primarily on appearance-based defects~\cite{8954181}, such as scratches or cracks, where anomalies manifest in static images. However, many real-world failures arise from violations of physical laws: a bearing that rotates irregularly or a gear system that skips motion cycles. Detecting such physics-grounded anomalies requires reasoning about dynamics and causal relations, not just appearance. This issue was introduced by Phys-AD~\cite{li2025towards}, the first dataset in this field, that features clear physical information and standardized content. Hence, our method is mainly developed on this dataset.

Recent advances in Vision-Language Models (VLMs) have unlocked impressive general-purpose reasoning ability. Yet, despite their broad pre-training, they lack explicit grounding in physics. As shown in the Phys-AD benchmark, current state-of-the-art models achieve only 66.9\% AUROC, underscoring their inability to distinguish between physically correct and anomalous motions. The gap arises because large-scale pre-training endows VLMs with semantic correlations, but not the causal structure of dynamics.

We address this challenge with a novel \textbf{physics-informed instruction fine-tuning} strategy. Instead of relying solely on video data, we inject explicit physical priors into the VLM through structured instructional prompts. These instructions, covering object identity, dynamic parts, and normative motion, are delivered in a step-wise multi-turn dialogue. By framing physics knowledge as a conversational learning process, the VLM is guided to form causal representations that generalize to unseen anomalies.

\begin{figure}[t!]
  \centering
  \centerline{\includegraphics[width=8.5cm]{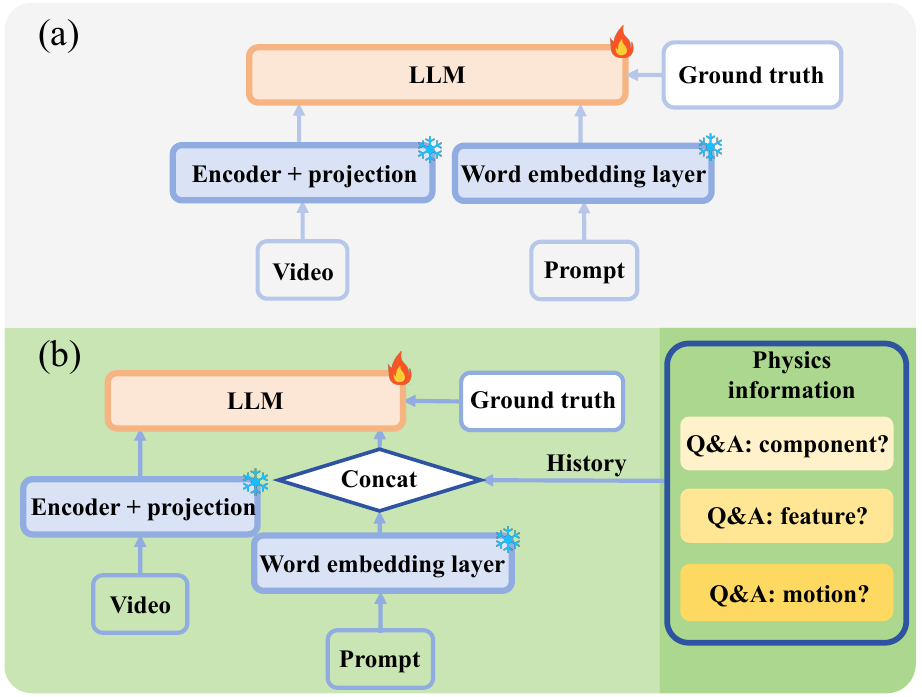}}
\caption{\textbf{Comparison of (\textbf{a}) conventional instruction fine-tuning and (\textbf{b}) our multi-turn physics-informed approach.} Our method uses multi-turn dialogue to incorporate structured physical information as prior knowledge, enhancing the LLM's autoregressive generation with improved logic and structure. This universal physical prior reduces dataset construction burden.}
\label{fig:compare}
\end{figure}

As Figure \ref{fig:compare}(a) shows, conventional instruction fine-tuning uses a prompt with a video, supervised by ground truth. By contrast, (b) shows that our approach adopts multi-turn dialogue, introducing physical information as prior knowledge. This approach endows the knowledge with better logic and structure, providing a solid basis for autoregressive LLM generation. Additionally, the universality of physical information makes prior knowledge generic within each object category, avoiding heavy burden of the dataset construction.

Our main contributions are:
\begin{itemize}
    \item We introduce an instruction fine-tuning strategy that explicitly encodes physics priors as structured, domain-specific instructions for VLMs.
    \item We design a multi-turn dialogue mechanism that decomposes physical reasoning into incremental steps, facilitating robust reasoning.
    \item We demonstrate state-of-the-art results on the Phys-AD benchmark, with large gains in both anomaly detection accuracy (96.7\% AUROC) and causal explanation quality (0.777 LLM score).
\end{itemize}

\section{Related Work}
\label{sec:related_work}

\noindent\textbf{Appearance-based vs. Dynamic Anomaly Detection}.
Traditional anomaly detection has focused on appearance-based defects, both in industrial settings~\cite{8954181, roth2022towards} and general video contexts~\cite{sultani2018real}. Early deep learning video methods relied on reconstruction~\cite{gong2019memorizing, park2020learning, lv2021learning} or pseudo-labels~\cite{wu2022self}, targeting visual novelty. The introduction of CLIP~\cite{radford2021learning} spurred a new wave of vision-language methods~\cite{wu2024vadclip, huang2025multimodal}, which were later improved with techniques like learnable prompts~\cite{zhou2022learning} and embedding space adjustments~\cite{zhou2023anomalyclip}. Recently, LLMs have been used for anomaly detection and explanation via captioning~\cite{zanella2024harnessing}, description matching~\cite{gao2025suvad}, and advanced prompting~\cite{zhang2025towards, gu2024anomalygpt}.
Despite these advances, the core limitation remains: these methods lack explicit physical grounding. The Phys-AD benchmark~\cite{li2025towards} formalized this gap, demonstrating the failure of existing models on kinematic irregularities. Our work directly addresses this by teaching the underlying physical laws.

\vspace{0.5mm}
\noindent\textbf{Instruction Tuning of Vision-Language Models}.
Instruction fine-tuning is the standard paradigm for aligning large models to user intent~\cite{lin2023video, zhang2023video}, though typically for general-purpose skills. Our work innovates by repurposing it for \textbf{deep knowledge injection}. By creating a specialized instruction set from physics principles, we demonstrate that this paradigm can instill domain-specific, causal expertise into a generalist VLM, moving beyond simple task-following to knowledge-based reasoning.

\begin{figure}[t!]
  \centering
  \centerline{\includegraphics[width=7.5cm]{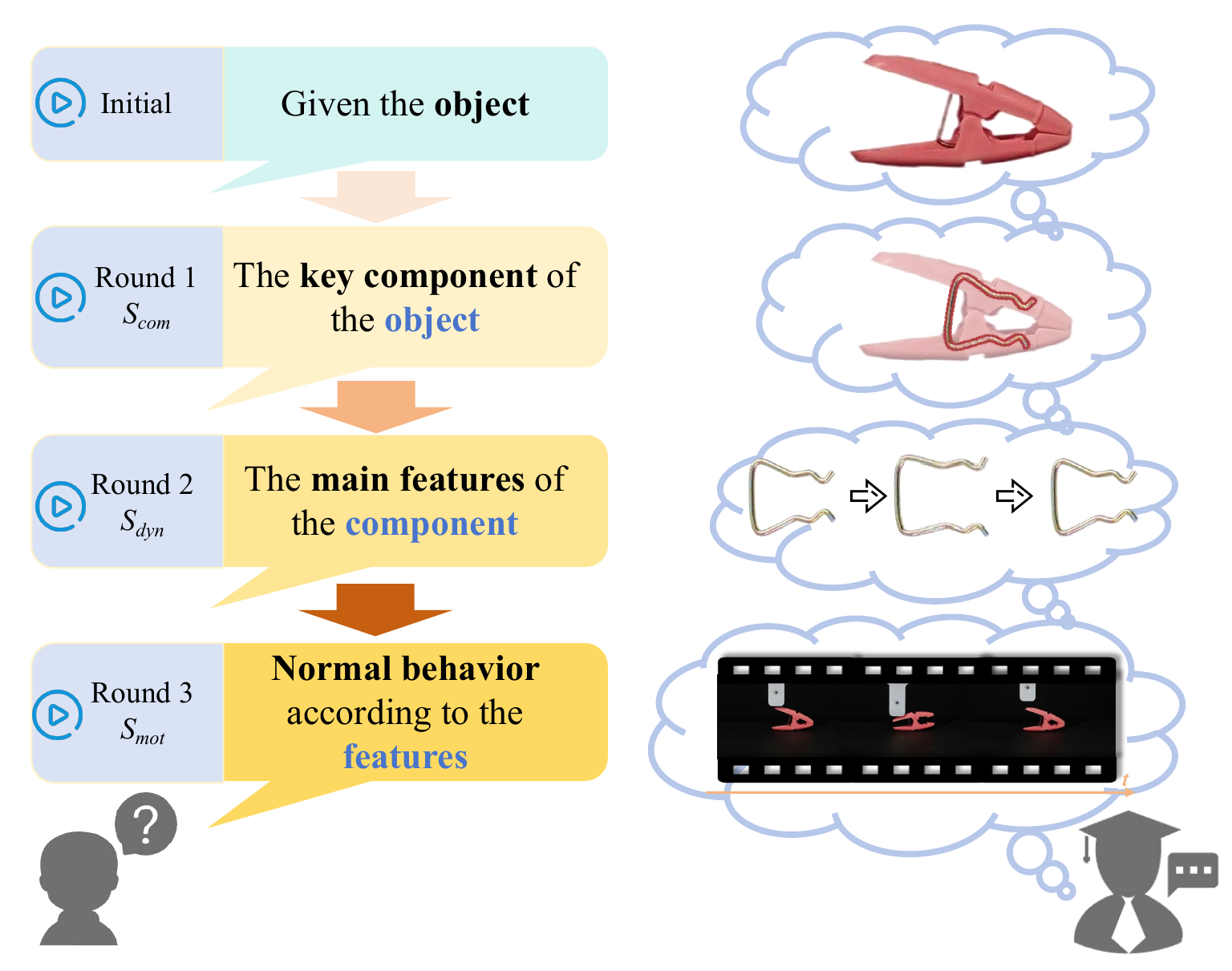}}
  \vspace{-0.4cm}
\caption{\textbf{The construction process of our structured physics information.} This explicit knowledge base, $\mathcal{P}_c=(S_{com}, S_{dyn}, S_{mot})$, guides the model through a logical reasoning chain for physics-grounded anomaly detection.}
\label{fig:rule}\vspace{-1mm}
\end{figure}

\section{Methodology}
\label{sec:methodology}

Our approach grounds a Vision-Language Model (VLM) in physical principles using an instruction fine-tuning strategy. We first formalize the problem of physics-grounded anomaly detection and then introduce our two core contributions: (1) a structured framework for representing physical knowledge as explicit instructions, and (2) a multi-turn dialogue mechanism to effectively inject this knowledge into the VLM.

\subsection{Problem Formulation}
\label{ssec:problem_formulation}

Given a video $V = \{f_1, f_2, \dots, f_T\}$ composed of $T$ frames, the task is to learn a function $\mathcal{F}$ that maps $V$ to an output pair $(y, E)$. Here, $y \in \{0, 1\}$ is a binary label indicating a \textit{normal} ($y=0$) or \textit{anomalous} ($y=1$) motion, and $E$ is a natural language string providing a causal explanation for the verdict. We aim to train a VLM, parameterized by $\theta$, to approximate this function:
\begin{equation}
    \mathcal{F}(V; \theta) \mapsto (y, E)
\end{equation}
The central challenge is that $\mathcal{F}$ must evaluate the video against implicit, category-specific physical laws not present in standard pre-training data.

\begin{figure*}[t!]
  \centering
  \centerline{\includegraphics[width=0.85\textwidth]{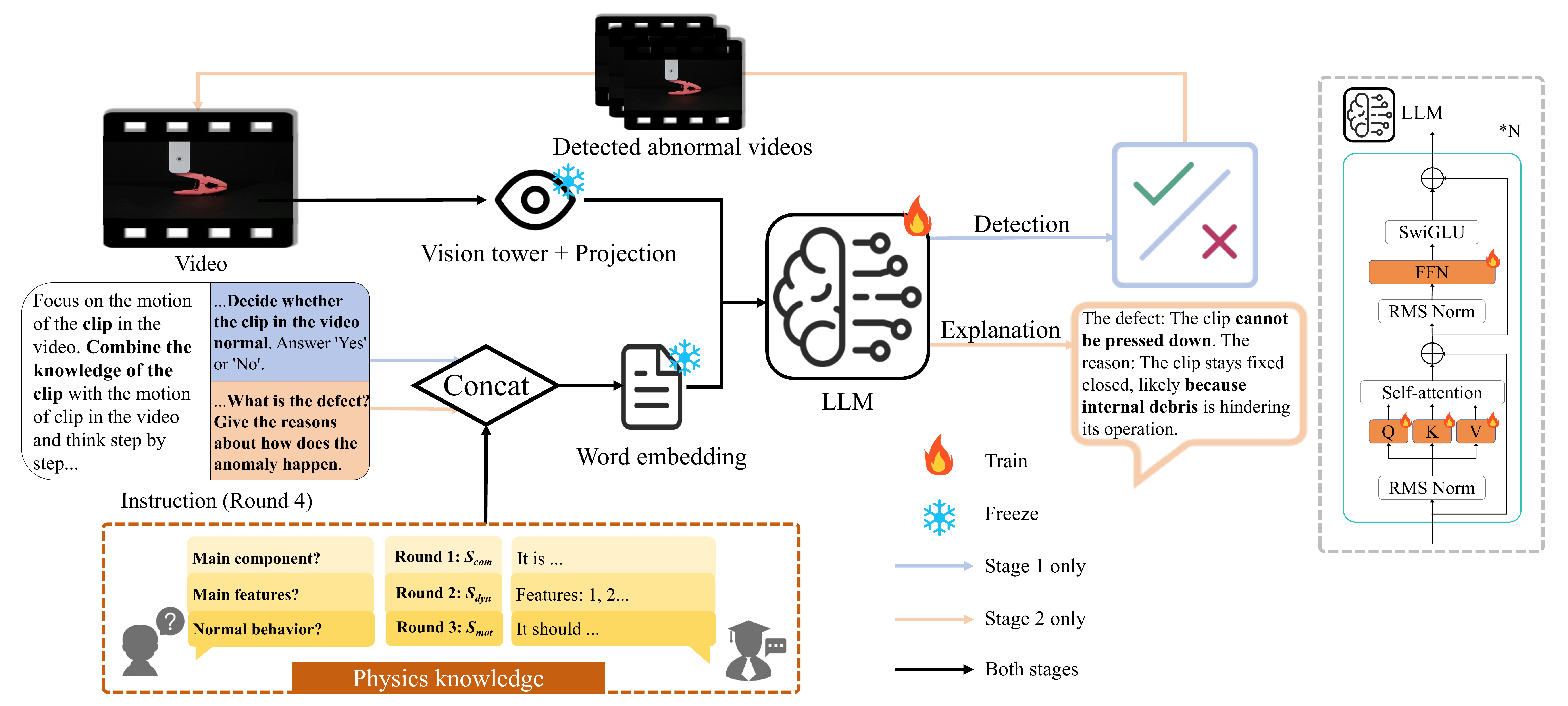}}\vspace{-3mm}
\caption{\textbf{Our model architecture for physics-grounded anomaly detection and explanation.} Based on Video-LLaVA, only the LLM components (QKV matrices and FFN layer) are fine-tuned, while the vision tower and multi-modal projector remain frozen. The multi-turn dialogue injects physical priors, enabling robust anomaly detection and causal explanation in two stages.}
\label{fig:pipeline}\vspace{-1mm}
\end{figure*}

\subsection{Structuring Physics as Textual Priors}
\label{ssec:method_formulation}

Our solution makes these implicit physical principles explicit. The cornerstone of our method is the formalization of abstract physical knowledge into a structured, machine-readable format. For each object category $c$, we define a \textbf{physics prior tuple}, $\mathcal{P}_c = (S_{com}, S_{dyn}, S_{mot})$. As illustrated in Figure~\ref{fig:rule}, this tuple serves as a knowledge base, with each component guiding the model through a logical reasoning step:

\noindent\textbf{1) Component Recognition ($S_{com}$):} Grounds the model by identifying the key component of the object and linking it to pre-existing semantic knowledge (e.g., "The clip has a spring mechanism, which possesses elasticity.").

\noindent\textbf{2) Dynamic Subject Focusing ($S_{dyn}$):} Deconstructs the system’s kinematics by identifying the principal moving parts and their physical properties (e.g., `The spring should deform under force and rebound upon release.').

\noindent\textbf{3) Motion Specification ($S_{mot}$):} Defines the observable motion of the object based on the subject's properties, establishing the expected spatio-temporal behavior (e.g., `Therefore, the clip's jaws should open when pressed and close when released.').

\subsection{Learning via Multi-Turn Dialogue}

We inject the structured knowledge from $\mathcal{P}_c$ into the VLM using a multi-turn instructional dialogue. For each training sample $(V, y, E)$, the model processes a sequence of four prompt-response pairs, $\{(Q_i, A_i)\}_{i=1}^4$.

The dialogue first instills the normative physical model before asking for a diagnosis. The target responses for the first three turns are set to the components of the physics prior, forcing the model to internalize them:
$A_1 = S_{com}$, $A_2 = S_{dyn}$, and $A_3 = S_{mot}$.

The final turn transitions from knowledge assimilation to diagnostic reasoning. The target response $A_4$ combines the ground-truth verdict and explanation, formatted as $\phi(y, E)$ (e.g., `Verdict: Anomalous. Reason: [E]').

The model parameters $\theta$ are optimized by minimizing the standard auto-regressive loss over the dialogue sequence:
\begin{equation}
    \mathcal{L}(\theta) = - \sum_{i=1}^{4} \log P(A_i | V, Q_1, A_1, \dots, Q_i; \theta)
    \label{eq:loss}
\end{equation}
This multi-turn structure is crucial. A naive single-turn approach, which concatenates all information into one response, suffers from \textbf{gradient dilution}: the repetitive physics text can dominate the loss signal, overshadowing the critical verdict information. Our step-by-step dialogue decomposes the task, ensuring the model first builds a robust knowledge foundation before applying it for diagnosis.

\subsection{Inference and Implementation}

\noindent\textbf{Inference.} At test time, a new video $V_{\text{new}}$ is presented with the same three-turn physics-grounding dialogue. The model then autoregressively generates the final answer $\hat{A}_4$, which is parsed to extract the predicted verdict $\hat{y}$ and explanation $\hat{E}$.

\noindent\textbf{Architecture.} We build on \textbf{Video-LLaVA}~\cite{lin2023video}, which uses a ViT vision tower and the Vicuna-7B LLM. As shown in Figure~\ref{fig:pipeline}, we follow standard efficient fine-tuning practices: the vision tower and projector are kept frozen, and we \textbf{only update the parameters of the LLM} (specifically, the self-attention and feed-forward layers) during training. This focuses the learning on integrating the new physics knowledge with the model's existing reasoning capabilities.

\section{Experiments}
\label{sec:experiments}

\subsection{Experimental Setup}

\noindent\textbf{Dataset and Model.} We evaluate our method on the \textbf{Phys-AD} benchmark~\cite{li2025towards}, which consists of 21 categories of objects with physics-grounded anomalies. We use \textbf{Video-LLaVA}~\cite{lin2023video} as our base VLM and train models on a single NVIDIA RTX 4090 GPU.

\noindent\textbf{Evaluation Metrics.} For anomaly detection, we report video-level Area Under the Receiver Operating Characteristic curve (AUROC). For anomaly explanation, we use two metrics: \textbf{SBERT Score} (cosine similarity between sentence embeddings of predicted and ground-truth explanations) and \textbf{LLM Score} (a refined PAEval metric~\cite{li2025towards}, where an external LLM, `deepseek-v3', scores predictions from 0 to 1 based on anomaly type correctness and reasoning logic).

\subsection{Results}

\noindent\textbf{Anomaly Detection.} As shown in Table \ref{tab:detection_results}, our physics-informed approach achieves an average AUROC of \textbf{96.7\%}, significantly outperforming prior methods, including the previous VLM-based SOTA (LAVAD at 51.0\%). This demonstrates the superior efficacy of explicitly teaching physical principles over implicit inference. {Our method achieves perfect or near-perfect scores on 14 of 21 categories}.

\vspace{0.5mm}
\noindent\textbf{Anomaly Explanation.} Table \ref{tab:explanation_results} presents causal explanation results. Our model achieves an average SBERT score of 0.824 (high semantic relevance) and an average LLM score of \textbf{0.777}, a substantial improvement over Video-ChatGPT (0.261). This validates that our model generates causally and logically correct explanations, not merely relevant keywords.

\begin{table}[t!]
\centering
\caption{\textcolor{black}{\textbf{Video-level AUROC ($\uparrow$) result  on Phys-AD.} 
}}
\centering\setlength{\tabcolsep}{1mm}
\resizebox{0.48\textwidth}{!}{
\begin{tabular}{l|cccccc|c}
\toprule

\textbf{Category} & \makecell[c]{\textbf{LAVAD}} & \makecell[c]{\textbf{ZS Clip}} & \makecell[c]{\textbf{ZS ImageBind}} & \makecell[c]{\textbf{Video-ChatGPT}} & \makecell[c]{\textbf{Video-LLaMA}} & \makecell[c]{\textbf{Video-LLaVA}} &\makecell[c]{\textbf{Ours}} \\ \midrule

Car               &0.557 &0.500 &0.500 &0.500 &0.678&0.522 & \textbf{0.999}\\
Fan               &0.510 &0.500 &0.500 &0.549 &0.592&0.611 & \textbf{0.912}\\
Rolling Bearing   &0.532 &0.500 &0.500 &0.300 &0.500&0.500 & \textbf{1.000}\\
Spherical Bearing &0.435 &0.500 &0.500 &0.450 &0.550&0.500 & \textbf{1.000}\\
Servo             &0.502 &0.500 &0.500 &0.506 &0.683&0.464 & \textbf{0.926}\\
Clip              &0.516 &0.500 &0.500 &0.669 &0.556&0.458 & \textbf{0.974}\\
USB               &0.513 &0.500 &0.500 &0.565 &0.575&0.500 & \textbf{0.964}\\
Hinge             &0.564 &0.500 &0.500 &0.500 &0.692&0.500 & \textbf{0.818}\\
Sticky Roller     &0.266 &0.500 &0.500 &0.450 &0.544&0.467 & \textbf{1.000}\\
Caster Wheel      &0.615 &0.500 &0.500 &0.444 &0.642&0.500 & \textbf{1.000}\\
Screw             &0.688 &0.500 &0.500 &0.472 &0.256&0.550 & \textbf{0.981}\\
Lock              &0.341 &0.500 &0.500 &0.279 &0.494&0.500 & \textbf{1.000}\\
Gear              &0.603 &0.500 &0.500 &0.544 &0.603&0.517 & \textbf{0.991}\\
Clock             &0.500 &0.500 &0.500 &0.501 &0.360&0.500 & \textbf{0.886}\\
Slide             &0.425 &0.500 &0.500 &0.562 &0.567&0.179 & \textbf{1.000}\\
Zipper            &0.535 &0.500 &0.500 &0.547 &0.489&0.500 & \textbf{0.945}\\
Button            &0.439 &0.500 &0.500 &0.515 &0.517&0.360 & \textbf{0.987}\\
Liquid            &0.504 &0.500 &0.500 &0.410 &0.278&0.217 & \textbf{1.000}\\
Rubber Band       &0.511 &0.500 &0.500 &0.517 &0.450&0.450 & \textbf{0.958}\\
Ball              &0.603 &0.500 &0.500 &0.562 &0.636&0.533 & \textbf{0.973}\\
Magnet            &0.502 &0.500 &0.500 &0.683 &0.300&0.400 & \textbf{1.000}\\ \midrule
Average           &0.510 &0.500 &0.500 &0.496 &0.523&0.463 & \textbf{0.967}\\ \bottomrule
\end{tabular}
}
\label{tab:detection_results}\vspace{-6mm}
\end{table}
\begin{table}[t!]
\centering
\caption{\textcolor{black}{\textbf{Video-level SBERT ($\uparrow$) result  on Phys-AD dataset.} 
}}
\vspace{-2mm}
\centering\setlength{\tabcolsep}{2mm}
\resizebox{0.48\textwidth}{!}{
\begin{tabular}{l|cc|cc|cc|cc}
\toprule
\multirow{2}{*}{\textbf{Category}} &
    \multicolumn{2}{c|}{\textbf{Video-ChatGPT}} &
    \multicolumn{2}{c|}{\textbf{Video-LLaMA}} &
    \multicolumn{2}{c|}{\textbf{Video-LLaVA}} &
    \multicolumn{2}{c}{\textbf{Ours}} \\
    \cmidrule(l){2-3}
    \cmidrule(l){4-5}
    \cmidrule(l){6-7}
    \cmidrule(l){8-9}
& \makecell[c]{SBERT} & \makecell[c]{LLM} & \makecell[c]{SBERT} & \makecell[c]{LLM} & \makecell[c]{SBERT} & \makecell[c]{LLM} & \makecell[c]{SBERT} & \makecell[c]{LLM}\\ \midrule
Car               &0.703&0.074  &0.577&0.108  &0.580&0.052  & \textbf{0.758}& \textbf{0.875}\\
Fan               &0.775&0.234  &0.715&0.557  &0.619&0.135  & \textbf{0.844}& \textbf{0.875}\\
Rolling Bearing   &\textbf{0.755}&0.330  &0.583&0.200  &0.729&0.268  & 0.746& \textbf{0.997}\\
Spherical Bearing &0.836&0.050  &0.634&0.158  &0.826&0.213  & \textbf{0.883}& \textbf{1.000}\\
Servo             &0.738&0.255  &0.642&0.304  &0.706&0.277  & \textbf{0.743}& \textbf{0.491}\\
Clip              &0.647&0.270  &0.517&0.083  &0.511&0.223  & \textbf{0.946}& \textbf{0.983}\\
USB               &0.766&0.164  &0.691&0.098  &0.732&0.171  & \textbf{0.837}& \textbf{0.887}\\
Hinge             &0.720&0.480  &0.675&0.377  &0.614&0.230  & \textbf{0.711}& \textbf{0.565}\\
Sticky Roller     &0.805&0.240  &0.708&0.220  &0.676&0.110  & \textbf{0.850}& \textbf{0.513}\\
Caster Wheel      &0.704&0.135  &0.662&0.156  &0.574&0.000  & \textbf{0.764}& \textbf{0.863}\\
Screw             &0.656&0.283  &0.613&0.314  &0.485&0.008  & \textbf{0.995}& \textbf{0.933}\\
Lock              &0.804&0.459  &0.646&0.351  &0.756&0.394  & \textbf{0.826}& \textbf{0.910}\\
Gear              &0.720&0.264  &0.679&0.415  &0.602&0.169  & \textbf{0.825}& \textbf{0.888}\\
Clock             &0.720&0.208  &0.607&0.161  &0.537&0.093  & \textbf{0.732}& \textbf{0.729}\\
Slide             &0.775&0.262  &0.663&0.201  &0.701&0.217  & \textbf{0.894}& \textbf{0.561}\\
Zipper            &0.555&0.549  &0.551&0.599  &0.429&0.322  & \textbf{0.782}& \textbf{0.951}\\
Button            &0.618&0.267  &0.539&0.254  &0.435&0.034  & \textbf{0.869}& \textbf{0.846}\\
Liquid            &0.686&0.339  &0.558&0.095  &0.599&0.136  & \textbf{0.752}& \textbf{0.384}\\
Rubber Band       &0.744&0.338  &0.567&0.210  &0.596&0.167  & \textbf{0.884}& \textbf{0.940}\\
Ball              &0.648&0.157  &0.587&0.044  &0.488&0.004  & \textbf{0.806}& \textbf{0.881}\\
Magnet            &0.655&0.130  &0.581&0.070  &0.459&0.013  & \textbf{0.855}& \textbf{0.363}\\ \midrule
Average           &0.716&0.261  &0.619&0.237  &0.603&0.154  & \textbf{0.824}& \textbf{0.777}\\ \bottomrule
\end{tabular}
}
\label{tab:explanation_results}\vspace{-3mm}
\end{table}

\begin{figure}[t!]
  \centering\vspace{5mm}
  \centerline{\includegraphics[width=0.48\textwidth]{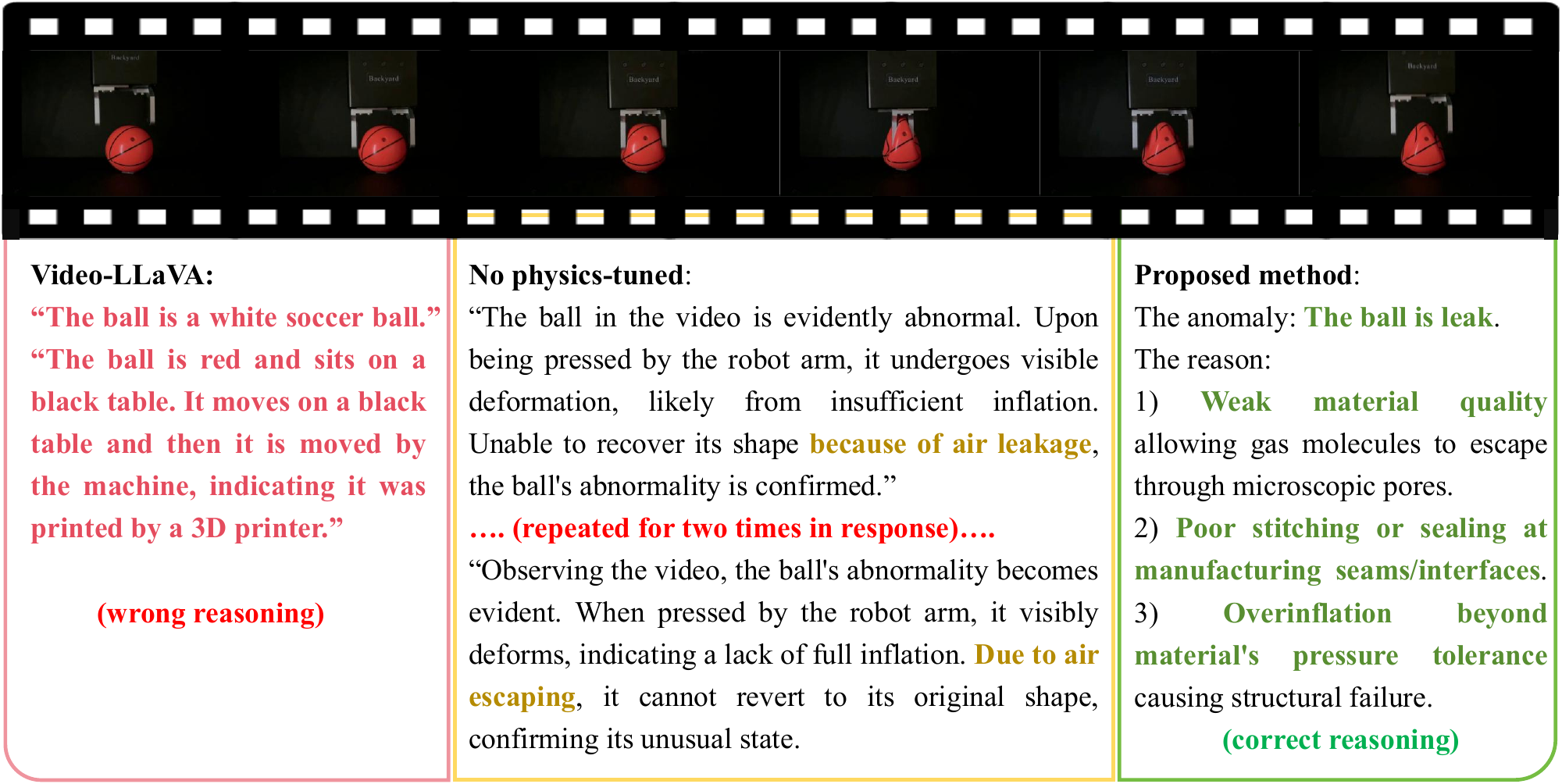}}\vspace{-3mm}
\caption{\textbf{Qualitative comparison of anomaly explanation methods on a `leaking ball' scene.} Our multi-turn physics-informed VLM provides a more accurate and causally relevant explanation than prior SOTA methods.}
\label{fig:conversation}\vspace{-3mm}
\end{figure}

\vspace{0.5mm}
\noindent\textbf{Qualitative Results.} Figure \ref{fig:conversation} illustrates that prior anomaly explanation methods often yield irrelevant or misinterpreted results. While standard fine-tuning improves relevance, it produces repetitive causal explanations. Our physics-informed approach, however, enhances object comprehension, leading to more diverse and accurate reasoning.

\begin{table}[t!]
\centering
\caption{\textcolor{black}{\textbf{Detection performance comparison between single-turn and multi-turn conversation. Evaluated with video-level AUROC ($\uparrow$) result of 21 categories on Phys-AD dataset.} Failure cases (AUROC $\approx$ 0.5) are highlighted in \textcolor{red}{red}.}}
\vspace{5pt}
\centering\setlength{\tabcolsep}{1mm}
\resizebox{0.48\textwidth}{!}{
\begin{tabular}{l|ccccccccccc}
\toprule

& \makecell[c]{Car} &\makecell[c]{Fan} & \makecell[c]{Rolling Bearing} &\makecell[c]{Spherical Bearing}& \makecell[c]{Servo} &\makecell[c]{Clip}& \makecell[c]{USB} &\makecell[c]{Hinge}& \makecell[c]{Sticky Roller}&\makecell[c]{Ball} &\makecell[c]{Magnet} \\ \midrule

No-phys             &0.994 &0.912 &1.000 &1.000 &0.877 &1.000 &0.958 &\textcolor{red}{0.557} &1.000 &0.935 &1.000\\
Single-turn         &1.000 & 0.868 & 1.000 & 1.000 & 0.898 & 1.000 & 0.969 &\textcolor{red}{0.500} &\textcolor{red}{0.500}&0.919 &\textcolor{red}{0.500} \\
Multi-turn          &0.999 & 0.912 & 1.000 & 1.000 & 0.926 & 0.974 & 0.964 & 0.818 &1.000 & 0.973 & 1.000\\
\bottomrule
\end{tabular}
}
\\
\vspace{5pt}

\resizebox{0.48\textwidth}{!}{
\begin{tabular}{l|cccc cccc cc|c}
\toprule

&\makecell[c]{Caster Wheel} & \makecell[c]{Screw} &\makecell[c]{Lock}& \makecell[c]{Gear}  &\makecell[c]{Clock} &\makecell[c]{Slide} &\makecell[c]{Zipper} &\makecell[c]{Button} &\makecell[c]{Liquid}  &\makecell[c]{Rubber Band}  &\makecell[c]{Average}\\ \midrule

No-phys             &0.905 &0.846 &1.000 &0.986 &0.757 &1.000 &0.932 &1.000 &1.000 &0.771 &0.925\\
Single-turn         &\textcolor{red}{0.500} &\textcolor{red}{0.500} & 1.000 & 0.988 &\textcolor{red}{0.549} &\textcolor{red}{0.514} & 0.833 & 1.000 & 0.981 &\textcolor{red}{0.500}  &0.787 \\
Multi-turn          & 1.000 & 0.981 & 1.000 & 0.991 & 0.886 & 1.000 & 0.945 & 0.987 & 1.000 & 0.958  & 0.967\\
\bottomrule
\end{tabular}
}
\label{tab:ablation}\vspace{-1mm}
\end{table}
\subsection{Ablation Study}

To validate our multi-turn dialogue design, we compare it against two ablated baselines. The single-turn approach concatenates all three physics prior components with the final verdict into one large target response. As shown in Table \ref{tab:ablation}, the multi-turn approach significantly outperforms the 'No-phys' (standard SFT) baseline, especially on challenging categories. Notably, the single-turn method suffered from convergence failure on 8 of 21 categories, yielding performance equivalent to random guessing (AUROC $\approx$ 0.5), highlighted by {red} text. This confirms that a {step-by-step instructional process} {is vital} for proper assimilation of complex knowledge.

\section{Conclusion}
\label{sec:conclusion}
We introduced a physics-informed instruction tuning framework to address the critical gap in VLM reasoning for dynamic, physics-grounded anomaly detection. By formalizing physical principles as structured instructions and delivering them through a multi-turn dialogue, we successfully taught a general-purpose VLM to perform complex causal reasoning about dynamics. Our approach establishes a new state-of-the-art on the Phys-AD benchmark. This work paves the way for more reliable and capable AI systems, where understanding and reasoning about physical laws is paramount.

\vfill\pagebreak
\bibliographystyle{IEEEbib}
\bibliography{refs}

\end{document}